\pdfoutput=1

\documentclass[11pt]{article}

\usepackage[final]{acl}

\usepackage{times}
\usepackage{latexsym}

\usepackage[T1]{fontenc}

\usepackage[utf8]{inputenc}

\usepackage{microtype}

\usepackage{inconsolata}

\usepackage{graphicx}

\usepackage{longtable}

\usepackage{forest,adjustbox}
\useforestlibrary{linguistics}
\forestapplylibrarydefaults{linguistics}
\usepackage{avm}

\usepackage{booktabs}
\usepackage{multicol}
\usepackage{multirow}
\usepackage{listings}
\usepackage{fvextra}

\usepackage{subcaption}

\title{More Aligned, Less Diverse? \\ Analyzing the Grammar and Lexicon of Two Generations of LLMs}

\author{
\textbf{Adrián Gude}\textsuperscript{1} \quad
\textbf{Roi Santos-Ríos}\textsuperscript{1$\ast$} \quad
\textbf{Francis Bond}\textsuperscript{3} \quad
\textbf{Dan Flickinger}\textsuperscript{2} \\
\textbf{Carlos Gómez-Rodríguez}\textsuperscript{1} \quad
\textbf{Olga Zamaraeva}\textsuperscript{1} \\[4pt]
\texttt{\{adrian.lopez.gude, roi.santos.rios, carlos.gomez, olga.zamaraeva\}@udc.es} \\
\texttt{danflick@alumni.stanford.edu} \quad \texttt{francis.bond@upol.cz} \\[6pt]
\normalsize
\textsuperscript{1}Universidade da Coruña, CITIC \quad
\textsuperscript{2}Independent Researcher \quad
\textsuperscript{3}Palacký University, Olomouc
}

\begin{document}
\maketitle
    \begin{abstract}

        This study contributes to a growing line of research in comparing LLM-generated texts with human-authored text, in this case, English news text. We focus in particular on the evaluation of syntactic properties through formal grammar frameworks. Our  analysis compares two generations of LLMs in the context of two human-authored English news datasets from two different years. Employing the Head-Driven Phrase Structure Grammar (HPSG) formalism, we investigate the distributions of syntactic structures and lexical types of 
        AI-generated
        texts and contrast them with the corresponding distributions in the human-authored New York Times (NYT) articles. 
        We use diversity metrics from ecology and information theory to quantify variation in grammatical constructions and lexical types.
        We show that English news text has changed little in the given time frame, while newer LLMs display reduced syntactic and, especially, lexical diversity compared to older, non-instruction-tuned models. These findings point to future work in studying effects of instruction tuning, which, while enhancing coherence and adherence to prompts, may narrow the expressive range of model output.
    \end{abstract}

    \section{Introduction}


        Large language models (LLMs) are increasingly compared to human writers across a growing range of linguistic and stylistic dimensions (e.g., \citealt{reinhart2025llms,moon2025homogenizing,rashid2024humanizing}). However, it remains unclear how such comparisons should be made and which dimensions best capture the  differences. This limits our ability to draw robust conclusions about what makes human and machine writing distinct.
        
        In this paper,\footnote{Adrián Gude and Roi Santos-Ríos contributed equally to this work. Both authors should be regarded as joint first authors.} we take a step toward addressing this gap by examining diversity, both lexical and syntactic, as a consistently informative dimension for comparing human and LLM-generated text. Building on the formal grammatical framework of Head-Driven Phrase Structure Grammar (HPSG) and the English Resource Grammar (ERG) \citep{Ben:Fli:Oep:02,Flickinger:11}, we analyze variation in syntactic constructions 
        and lexical types using diversity metrics drawn from ecology and information theory (Shannon and Simpson indices; \citealt{Magurran2004,stamatatos2009survey}).
        
        In contrast to prior related work \citep{zamaraeva2025comparing,munoz2024contrasting}, we compare two generations of LLMs and human-authored news writing. For the human side, we analyze New York Times (NYT) lead paragraphs from two distinct periods (2023 and 2025). For the LLM side, we compare a suite of base models trained prior to 2023 (LLaMA, Mistral, Falcon) with newer instruction-tuned models trained up to 2024 (Qwen 2.5, Mistral 7B v0.3, GPT-4o, LLaMa 3.3).
        
        Our results show that English news text remains stable across time in all diversity metrics, suggesting a consistent balance of grammatical and lexical variety in professional news prose. 
        In contrast, in LLM-generated text, we observe that both syntactic and
        lexical diversity decline substantially in newer, instruction-tuned models, with the effect being especially pronounced for lexical diversity. In addition, the newer LLM texts are easier to parse and take less computer memory to do so.
        Our findings suggest that, while instruction tuning is designed to improve the helpfulness and coherence of responses to natural-language prompts \citep{NEURIPS2022_b1efde53}, it has the side effect of reducing the syntactic and lexical breadth of the outputs.  Instruction tuned models generate outputs that are stylistically narrower and less varied than both humans and, interestingly, than earlier base models.
        
        Overall, our study highlights diversity metrics as a robust, linguistically grounded way to track stylistic and grammatical shifts across model generations, shedding light on how current tuning paradigms may trade off lexical variety  for stylistic control.\footnote{Code available at: \url{https://github.com/olzama/llm-syntax/}}

    \section{Related work}
    \label{sec:related}
    
    \citet{munoz2024contrasting} conducted a large-scale quantitative analysis contrasting texts generated by base (non-instruction tuned) LLMs with human-written news texts. Their results revealed differences across multiple linguistic dimensions, including morphological, syntactic, psychometric, and sociolinguistic aspects. These findings established a detailed baseline highlighting how human linguistic patterns remain more diverse and less homogenized compared to model-generated text.

    \citet{zamaraeva2025comparing} further extended this line of research by conducting a  comparison on the same data using Head-Driven Phrase Structure Grammar (HPSG) and the English Resource Grammar (ERG), providing a fine-grained  perspective within an independent linguistic-theoretic framework. Their study showed that human-authored texts had greater grammatical but \emph{less} lexical diversity than the LLM texts, and that human writers differed from each other more than each differed with respect to any LLM (in other words, that LLMs act as an ``average'' writer).
    
    So far, little research has been done on the effects of instruction tuning and reinforcement learning from human feedback (RLHF) on LLMs' grammatical diversity. \citet{padmakumar2024doeswritinglanguagemodelS} found that RLHF affected vocabulary type/token ratios of LLMs across LLMs, leading to more homogeneous texts. Another  evaluation of the different stages of RLHF training showed that although RLHF improves out-of-distribution generalization compared to supervised fine-tuning, it significantly reduces output diversity measured through a combination of N-gram counting, semantic cosine similarity, and natural language inference metrics, revealing a tradeoff between adaptability and linguistic variety \citep{kirk2024understandingeffectsrlhfllm, shypula2025evaluating}. To summarize, prior research suggests that, while modern LLMs have improved in fluency and instruction-following capabilities, these advances may come at the cost of lexical and stylistic diversity even in the new families of LLMs. Human texts, by contrast, still exhibit greater variety and more complex linguistic structures. 

    Our study contributes to this idea in several key ways. First, while previous work has compared a single generation of LLMs to human text or analyzed the effects of RLHF in isolation, we conduct a direct, diachronic comparison between two generations of LLMs: older base models and newer, instruction-tuned ones. Second, we mirror this generational approach on the human side by contrasting these models against human-authored news texts from two corresponding time periods. Furthermore, we use a linguistic-theoretic grammar framework, namely HPSG, to provide a way of analyzing syntactic and lexical properties of language that is independent from natural language processing tasks and thus should be more robust/generalizable. This framework allows us to look at lexical distributions in a systematic manner, beyond the surface information that vocabulary counts provide. HPSG lexical types are complex representations of word types that specify aspects of their syntactic behavior. We are not aware of a similar resource within, e.g., the UD framework; POS distinctions are too coarse. Last but not least, we employ independent diversity metrics from ecology.
    

    \section{Methodology}
    \label{sec:methodology}

        We use the HPSG-based framework utilized e.g.\ by ~\citet{zamaraeva2025comparing} and focus on diversity metrics from ecology and information theory to quantify changes in grammatical and lexical variability over time.
        
        \subsection{English Resource Grammar}
        \label{sec:erg}
        
            The goal of this study is to provide an insight into how LLMs change over time with respect to the grammatical properties of their writing. For this purpose, it is not enough to look at just the vocabulary, and, while looking at dependency structures like Universal Dependencies \citep{nivre2016universal}, as \citet{munoz2024contrasting} did is useful, we are interested in a comparison within a framework that is rooted in a formal linguistic theory and not inherently biased towards performance on NLP tasks. For this reason, we choose the DELPH-IN HPSG framework, following \citet{zamaraeva2025comparing}. Head-driven Phrase Structure Grammar \citep[HPSG; ][]{Pol:Sag:94} is a theory of syntax that was developed by linguists theoretically and with empirical validation in mind, which is why the theory is associated with fully explicit formalisms that can be fully implemented on the computer, allowing for rigorous validation of the theoretical claims made about syntax. DELPH-IN\footnote{\url{https://delph-in.github.io/docs/home/Home/}} is one such formalism that matured into a long-term international grammar engineering effort. In particular, the English Resource Grammar \citep[ERG; ][]{flickinger2000building, Flickinger:11} has been in continuous development with regular releases,\footnote{\url{https://github.com/delph-in/erg/releases/tag/2025}} reaching 94\% average accuracy over a variety of English corpora. Importantly and in contrast to statistical parsers, the ERG is designed to provide structure for \emph{all possible well-formed} English sentences and to \emph{reject} strings that do not correspond to well-formed English utterances. This is crucial for our methodology, because we want to be able to compare LLMs between each other and to human writers with respect to \emph{rare} linguistic phenomena, not only the frequent ones, and also with respect to sentences which were \emph{not} parsed by the grammar for some reason that can be informative. 
            The second property of the HPSG/ERG that is important to us is its structure: the grammar is represented as a clear hierarchy of \emph{syntactic and lexical types}. The type definitions are detailed HPSG structures which specify sets of constraints which make a construction possible (such as: ``this is a verb and it requires two complements of which one is a noun'', to provide a simplistic example). Syntactic type definitions are used at parse time to ``license'' phrases bottom-up, until a complete sentence spanning the whole input string is built (such as noun phrases, verb phrases...), whereas lexical types are used by the parser only at the beginning of the parsing but provide very rich information about the specific constraints that need to be met. In this sense, lexical types are drastically different from vocabulary items which are just surface strings representing words. Finally, grammar-based exhaustive chart parsing provides us with a window into the sentences' \emph{parsability}: how easy or difficult is it to parse a particular sentence, as a proxy measure of how complex it is.

        \subsection{Portability}

            While our experiments focus on English, the grammar-based approach itself can be used with any language (notably, also with low-resource languages). The HPSG framework is based on general linguistic theory and is not specific to English, and the tools used to develop these grammars are likewise language-independent. At the same time, each implemented grammar necessarily includes language-specific layers, which requires expert effort to develop. As a result, existing resources differ in their size and coverage, with English currently having the most mature and broad-coverage grammars. However, the required investment is comparable to that of training machine learning systems, which depend on significant data and computational resources that may not be equally available across languages or domains.
            We therefore see the methodology as broadly applicable, with its extension to other languages primarily dependent on the continued development of high-quality grammatical resources. The same applies across different genres, as once an adequate grammar is available for a language, the approach can be readily transferred to non-news domains with not much additional effort.

        \subsection{Diversity metrics}
    
            We use two diversity measures,  the Shannon-Wiener diversity index (H, or Shannon Index), and the Simpson Diversity Index \citep{Magurran2004}.
            
            Shannon: $H' = - \sum_{i=1}^{S} p_i \ln(p_i)$
            
            Simpson: $D = 1- \sum_{i=1}^{S} p_i^2$

            The Shannon Index is widely used as an ecological measure of species diversity \citep{https://doi.org/10.1046/j.1466-822X.2003.00015.x}. It considers both the number of species (richness) and the evenness of their distribution, meaning a higher H value indicates greater diversity. The index is the same as the Shannon Entropy, and quantifies the uncertainty or "surprise" of predicting the next species in a community.  
            
            The Simpson Index \citep{Simpson1949} measures the probability that two individuals (or tokens) randomly selected from a sample will belong to different categories (e.g., species, construction, lexical type, \dots{}). It is less sensitive to low frequency phenomena than the Shannon Index. 
            


            \begin{table*}[ht!]
                \centering
                    \caption{Datasets: reproduced in full from Table 1 in \citealt{munoz2024contrasting}, alongside the experiments done in \citealt{zamaraeva2025comparing}.}
            
                \begin{adjustbox}{max width=\textwidth}
                    \begin{tabular}{lrrrl}
                        \toprule
                        Dataset & \# Sent. in dataset & Model size & Training tokens & Data sources \\
                        \midrule
                        {\multirow{4}{*}{{LLaMa}}} 
                        & 37,825 & 7B  & 1T   & \multirow{4}{*}{Not disclosed} \\
                        & 37,800 & 13B & 1T   & \\
                        & 37,568 & 30B & 1.5T & \\
                        & 38,107 & 65B & 1.5T & \\
                        \midrule
                        {\multirow{3}{*}{{Falcon}}} & & & & RefinedWeb-English (76\%), RefinedWeb-Euro (8\%), \\
                        & 27,769 & 7B & 1.5T & Gutenberg (6\%), Conversations (5\%) \\
                        & & & & GitHub (3\%), Technical (2\%) \\
                        \midrule
                        Mistral & 35,086 & 7B &  Not disclosed & Not disclosed \\
                        \midrule
                        Original NYT & 26,102 & N/A & N/A & New York Times Archive, Oct.\ 1, 2023 - Jan.\ 24, 2024 \\ 
                        \midrule
                        Redwoods (WSJ) & 43,043 & N/A & N/A & Wall Street Journal sections 1-21 \\ \cline{2-5}
                        Redwoods (Wikipedia) & 10,726 & N/A & N/A & Wikipedia \\
                        \bottomrule
                    \end{tabular}
                    \label{tab:dataset-sizes}
                    \end{adjustbox}
            \end{table*}
            
            \begin{table*}[ht!]
                \centering
                    \caption{Datasets contributed with this paper.}
            
                \begin{adjustbox}{max width=\textwidth}
                    \begin{tabular}{lrrrl}
                        \toprule
                        Dataset & \# Sent. in dataset & Model size & Training tokens & Data sources \\
                        \midrule
                         {\multirow{3}{*}{{Qwen 2.5}}} & 37,825 & 14B & 18T &  \\
                          & 26,498 & 32B & 18T & Not disclosed \\
                         & 34,892 & 72B & 18T & \\
                        \midrule
                        {\multirow{3}{*}{{LlaMa 3.3}}} 
                        &  &  &  & \multirow{3}{*}{Not disclosed} \\
                        & 39,306 & 70B & 15T+ & \\
                        &  &  &  & \\
                        \midrule Mistral v.0.3 & 33,840 & 7B & 1T & Not disclosed\\
                        \midrule 
                        GPT-4o & 50,544 & Not disclosed & 13T & Not disclosed\\
                        \midrule 
                        Original NYT & 26,102 & N/A & N/A & New York Times Archive, Feb.\ 1, 2025 - May.\ 31, 2025 \\ 
              
                        \bottomrule
                        
                    \end{tabular}
                    \label{tab:dataset-sizes-2}
                    \end{adjustbox}
            \end{table*}

    \section{Data and generative models}
    \label{sec:data}
      
        To study the differences between LLM-generated and human-authored news texts, we construct a new dataset similar in structure to the one used in \citet{munoz2024contrasting} and \citet{zamaraeva2025comparing}. We use New York Times (NYT) lead paragraphs from February to May 2025. We assume that the LLMs under investigation, all trained on data up to 2024, could not have encountered these human-authored articles in training.  

            \lstdefinestyle{promptbox}{
              basicstyle=\ttfamily\small,
              columns=fullflexible,
              breaklines=true,
              frame=single,
              framerule=0.4pt,
              backgroundcolor=\color[gray]{0.98},
              rulecolor=\color[gray]{0.75},
              xleftmargin=1ex,
              xrightmargin=1ex,
              framesep=0.6ex,
            }
    
            \begin{figure}[t]
            \centering
              \captionsetup{type=figure}
              \begin{minipage}{\linewidth}
                \begin{lstlisting}[style=promptbox]
system_prompt: "You are a professional journalist specializing in writing news. Follow the given structure."

user_prompt: "You will write a news lead paragraph using the inputs below.
  Inputs
  Headline: {headline}
  LeadThreeWords: {lead_three_words}
  Requirements - Mandatory
  Write one paragraph of several sentences (more than one, e.g. two-three (2-3)); no title, no bullets.
  Output format: the paragraph only, no preamble or labels."
                \end{lstlisting}
              \end{minipage}
              \caption{Prompts used to generate news lead paragraphs from LLMs (system and user prompts).}
              \label{fig:prompt-example}
            \end{figure}
        
        Our new NYT dataset mirrors the structure of \citealt{munoz2024contrasting}. Human-authored texts consist of lead paragraphs downloaded from the NYT Archive API.\footnote{\url{https://developer.nytimes.com/docs/archive-product/1/overview}} For each headline, we prompted our set of LLMs with the headline and the first three words of the lead paragraph to generate synthetic leads. Whereas~\citet{munoz2024contrasting} used base models that could be prompted directly for text completion, the instruction-tuned models need a more explicit prompt in the form of instructions telling them to complete the paragraph, which is shown in Figure~\ref{fig:prompt-example}. The analyses we present refer exclusively to the human-written and LLM-generated lead paragraphs 

        
        The LLMs in \citet{munoz2024contrasting} included earlier systems (LLaMA, Falcon 7B, and Mistral 7B, all released prior to October 2023), while in this study we extend the design to more recent models: Qwen 2.5 (14B, 32B, 72B), LLaMA 3.3 (70B), Mistral 7B v0.3, and GPT-4o. Following \citet{munoz2024contrasting}, we continue to distinguish scaling effects within a single architecture (e.g.\ different Qwen sizes) from differences due to model families.\footnote{As a total, we performed 3 initial text generations with Qwen 2.5 32B to calibrate the outputs of the LLMs, then one execution per model, and lastly, another execution per model with the latest prompt, disclosed in Figure \ref{fig:prompt-example}. The total number of executions amounts to 17.} 
        
        Dataset and model properties are summarized in Tables~\ref{tab:dataset-sizes}-\ref{tab:dataset-sizes-2}. The hyperparameters used for all models are the following: temperature: 0.7, top\_p: 0.92, top\_k: 50, repetition\_penalty: 1.05, max\_new\_tokens: 1000, num\_return\_sequences: 1, num\_beams: 1. The average sentence length is in the range of 18-20 tokens for 2023 LLMs, while the newer 2025 models generate around 22-29 tokens, as shown in Table \ref{tab:unbinned}.
        

        \subsection{Scope}
                We deliberately restrict our experimental scope to a controlled news genre when comparing linguistic properties of LLM-generated and human-authored texts. This design choice follows prior work \citep{zamaraeva2025comparing, munoz2024contrasting} and reflects both methodological and conceptual considerations. First, genre may exert an influence on linguistic structure, outweighing individual author effects \cite{biber1991variation}; limiting genre variation therefore reduces confounds and allows clearer attribution of observed differences to generation source rather than discourse conventions. Second, focusing on one genre enables more reliable measurement of fine-grained linguistic properties, which may otherwise be obscured by cross-genre heterogeneity. Finally, while large-scale data generation across multiple genres would be desirable, it is computationally and financially costly, and may incentivize breadth at the expense of analytical depth. Our  scope  prioritizes internal validity and interpretability, providing a principled basis for future work to test the generality of these findings across genres and domains.
        
    \section{Results}
    In the following subsections we compare the grammar distributions in the human-authored and the LLM-generated datasets. The distributions were obtained with the English Resource Grammar (\S\ref{sec:erg}). We provide a list of the most distinctive syntactic and lexical types, as well as examples of where they are found in the data. However, the main point is that the datasets produced by people and by older and newer LLMs can be distinguished at the level of grammatical types \textbf{distributions}, taken as a statistical snapshot. Examples are meant to be illustrative but not necessarily explanatory. Any dataset can contain any instance of any syntactic or lexical type.
    \label{sec:results}

            \begin{table*}
                \centering
                \scriptsize
                \begin{tabular}{p{2.5cm} l p{8cm} p{2cm}}
                \toprule
                \textbf{Construction} & \textbf{Preferred by} & \textbf{Example Sentence} & \textbf{Constituent} \\
                \midrule
                Nominal head + preceding adjunct & LLM &
                    Phyllis Dalton, the Oscar-winning costume designer known for her meticulous work on historical epics, passed away at the age of 99. &
                    historical epics \\
                
                \midrule
                 
                Subordinate pred phrase from participial VP & LLM &
                    The Learjet was carrying a pediatric patient when it crashed, killing all on board. &
                    killing all on board \\
                
                \midrule
                 
                Head + following scope adjunct & LLM &
                    Speaker Kevin McCarthy began the final day before the shutdown, facing dim prospects of passing a funding measure. &
                    facing dim prospects \\
                
                \midrule
                 
                Bare NP & LLM &
                    Tariff strategy could harm economic growth and job creation. &
                    job creation \\
                
                 
                
                \midrule
                 
                Proper-name bare NP & Humans &
                    Midterm elections reshaped the balance of power in Washington. &
                    Washington \\
                
                \midrule
                 
                Bare NP from quantified daughter & Humans &
                    Flights raised concerns over those being returned to Honduras. &
                    those being returned to Honduras \\
                
                \midrule
                 
                Fragment NP & Humans &
                    Chinese couple assembled building blocks: graduate degrees and careers. &
                    graduate degrees and careers \\
                
                \midrule
                 
                Measure NP & Humans &
                    Near-miss incidents occurred over a few years. &
                    few years \\
                
                
                
                \bottomrule
                \end{tabular}
                
                \caption{Syntactic constructions contributing the most to statistical differences between 2025 LLMs and English news text.}
                \label{tab:syntrank}
            \end{table*}

            \begin{table*}
                \centering
                \scriptsize
                \begin{tabular}{p{3cm} l p{8cm} p{2cm}}
                \toprule
                \textbf{Lex types} & \textbf{Preferred by} & \textbf{Example Sentence} & \textbf{Constituent} \\
                \midrule
                
                Adjective (intersective, non-comparative) & LLM &
                    A look back at her career reveals a dedication to authenticity and detail that brought eras long past vividly to life on screen. &
                    past \\
                
                \midrule
                
                Transitive verb with NP complement & LLM &
                Critics argue this action undermines collective bargaining rights and could lead to increased tensions with federal employee unions. &
                undermines\\
                
                \midrule
                
                Count noun (lexical) & LLM &
                    A man was wounded and a nine-year-old boy was fatally shot during an incident in Newark, according to officials.&
                    man \\
                
                \midrule
                
                Count noun with non-specific reference & LLM &
                    The unrest has displaced thousands and raised concerns over the control of vital resources such as cobalt and copper. &
                    concerns \\
                
                
                
                \midrule
                
                Proper noun (generic) & Human &
                    Internal government reports have revealed that Reagan National Airport experienced multiple near-miss incidents over the past few years. &
                    Airport \\
                
                \midrule
                
                Adj (intersective, coordinated, generic) & Human &
                    A farm in rural Oregon has made headlines with an extraordinary offer: 40,000 pounds of fresh salmon, completely free of charge. &
                    40,000 \\
                
                \midrule
                
                Verb (PP + finite CP complement, implicative) & Human &
                    Critics said the president's characterization was not only inaccurate but also deeply hurtful to individuals with disabilities. &
                    said \\
                
                \midrule
                
                Proper noun (referential) & Human &
                    More than a dozen prosecutors at the Washington U.S. Attorney's Office have been dismissed, according to sources familiar with the matter. &
                    Washington \\
                
                
                
                \bottomrule
                \end{tabular}
               \caption{Lexical types contributing the most to statistical differences between 2025 LLMs and English news text}
                \label{tab:lexrank}
            \end{table*}

            \begin{table*}
                \centering
                \scriptsize
                \begin{tabular}{p{2.5cm} l p{7cm} p{3cm}}
                \toprule
                \textbf{Construction} & \textbf{Year} & \textbf{Example Sentence} & \textbf{Constituent} \\
                \midrule
                Nominal head + preceding adjunct & 2025 &
                    Phyllis Dalton, the Oscar-winning costume designer known for her meticulous work on historical epics, passed away at the age of 99. &
                    historical epics \\
                
                \midrule
                 
                Subordinate pred phrase from participial VP & 2025 &
                    The Learjet was carrying a pediatric patient when it crashed, killing all on board. &
                    killing all on board \\
                
                \midrule
                 
                Head + following scope adjunct & 2025 &
                    Speaker Kevin McCarthy began the final day before the shutdown, facing dim prospects of passing a funding measure. &
                    facing dim prospects \\
                
                \midrule
                 
                Bare NP & 2025 &
                    Tariff strategy could harm economic growth and job creation. &
                    job creation \\
                
                 
                    
                \midrule         
                Adjective–Headed Normal Construction & 2023 & “I am concerned about how they are using [the law],” Cardin said. & they \\
                \midrule
                Bare NP & 2023 & Senator Ben Cardin told Al-Monitor that he will oppose the release of the remaining \$650 million in military aid to Egypt. & Senator Ben Cardin \\
                \midrule
                Subordinated Predicative VP (Participial) & 2023 & It was 40 degrees outside, but people were lined up wearing shorts and flip-flops. & but people were lined up wearing shorts and flip-flops. \\
                \midrule
                Head–Adjective Small Clause (Predicative) & 2023 & When the Supreme Court returns on Monday from its summer recess, it will be without Justice Antonin Scalia. & Justice Antonin Scalia. \\
                \bottomrule
                \end{tabular}
                
                \caption{Syntactic constructions contributing the most to the statistical differences between the 2023 and 2025 LLMs.}
                \label{tab:llm_syntconst_side_by_side}
            \end{table*}
            

            \begin{table*}
                \centering
                \scriptsize
                \begin{tabular}{p{3cm} l p{7cm} p{2cm}}
                \toprule
                \textbf{Lex types} & \textbf{Year} & \textbf{Example Sentence} & \textbf{Constituent} \\
                \midrule
                
                Adjective (intersective, non-comparative) & 2025 &
                    A look back at her career reveals a dedication to authenticity and detail that brought eras long past vividly to life on screen. &
                    past \\
                
                \midrule
                
                Transitive verb with NP complement & 2025 &
                Critics argue this action undermines collective bargaining rights and could lead to increased tensions with federal employee unions. &
                undermines\\
                
                \midrule
                
                Count noun (lexical) & 2025 &
                    A man was wounded and a nine-year-old boy was fatally shot during an incident in Newark, according to officials.&
                    man \\
                
                \midrule
                
                Count noun with non-specific reference & 2025 &
                    The unrest has displaced thousands and raised concerns over the control of vital resources such as cobalt and copper. &
                    concerns \\
                
                
                
                \midrule
                Utterance particle & 2023 & I don't see any signs that inventories are excessive. & I \\
                \midrule
                Pronoun (personal, first person singular, it) & 2023 & “It’s not acceptable for a democracy.” & It \\
                \midrule
                Pronoun (personal, second person, you) & 2023 & “If you ran for president,” he wondered, “would you be able to win in Iowa?” & you \\
                \midrule
                Pronoun (personal, first person plural, we) & 2023 & There was a time when we could blame our problems on Bush, but no more. & we \\
                \bottomrule
                \end{tabular}
                
                \caption{Lexical types contributing the most to the statistical differences between the 2023 and 2025 LLMs.}
                \label{tab:llm_lextypes_side_by_side}
            \end{table*}


        \subsection{Syntactic types: Humans and LLMs}
        \label{sec:constr}
            

            \begin{figure*}[t]
              \centering
              \begin{minipage}[t]{0.48\textwidth}
                \centering
                \vtop{\null\hbox{\includegraphics[width=\linewidth]{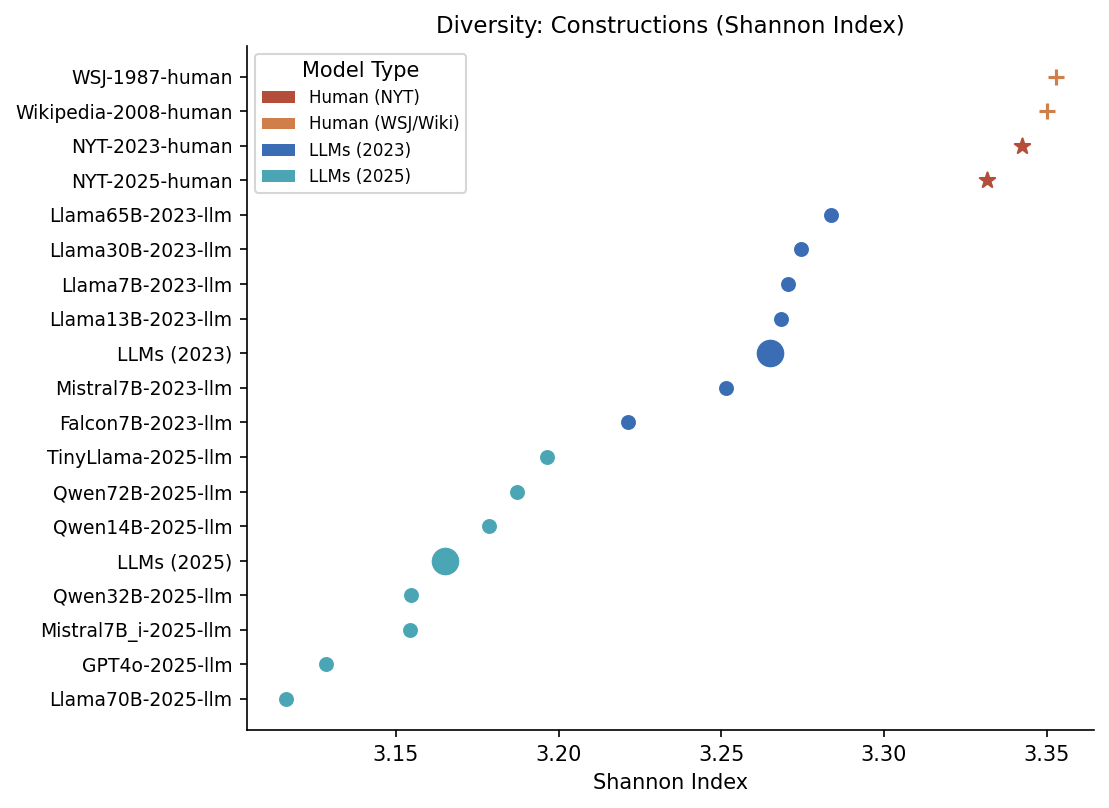}}}
                \caption{Syntactic construction diversity measured using the Shannon Index. Higher values indicate a more varied distribution of syntactic constructions. On the Y-axis, each point corresponds to a model name.}
                \label{fig:erg-const-shannon}
              \end{minipage}%
              \hfill
              \begin{minipage}[t]{0.48\textwidth}
                \centering
                \vtop{\null\hbox{\includegraphics[width=\linewidth]{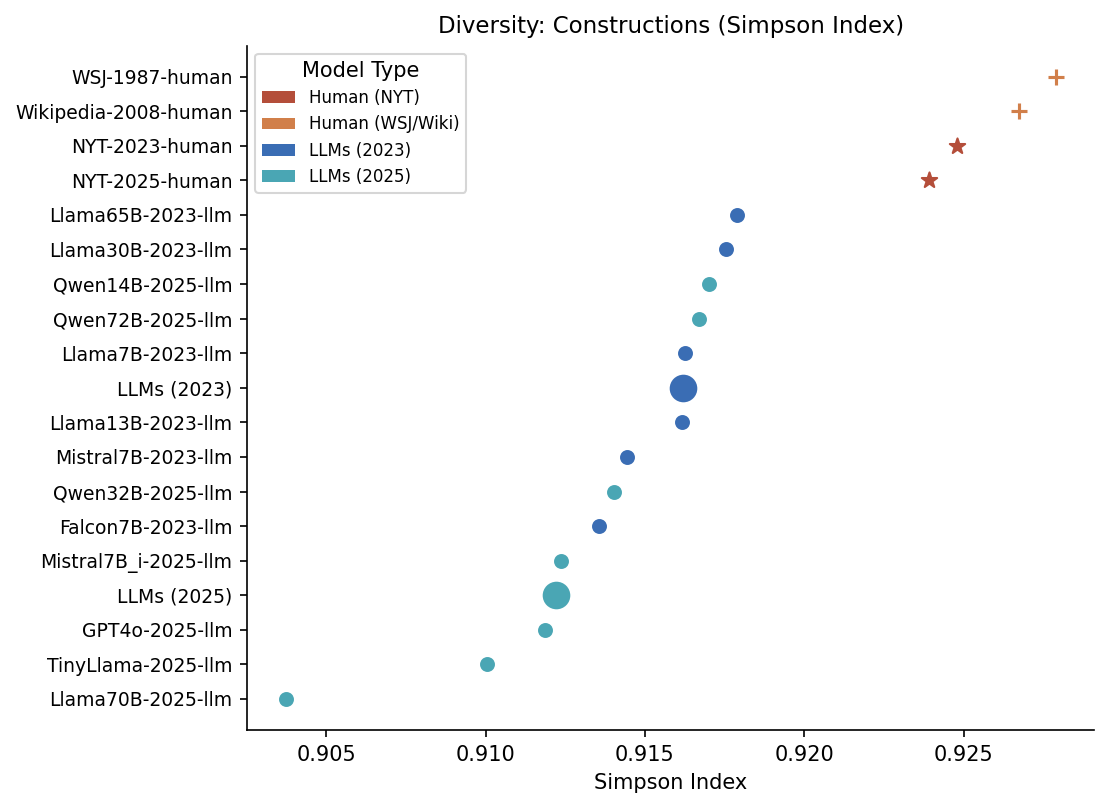}}}
                \caption{Syntactic construction diversity measured using the Simpson Index. Higher values indicate a more varied distribution of syntactic constructions. On the Y-axis, each point corresponds to a model name.}
                \label{fig:erg-const-simpson}
              \end{minipage}
            \end{figure*}

            \begin{figure*}[t]
              \centering
                \begin{minipage}[t]{0.48\textwidth}
                \centering                    \vtop{\null\hbox{\includegraphics[width=\linewidth,clip]{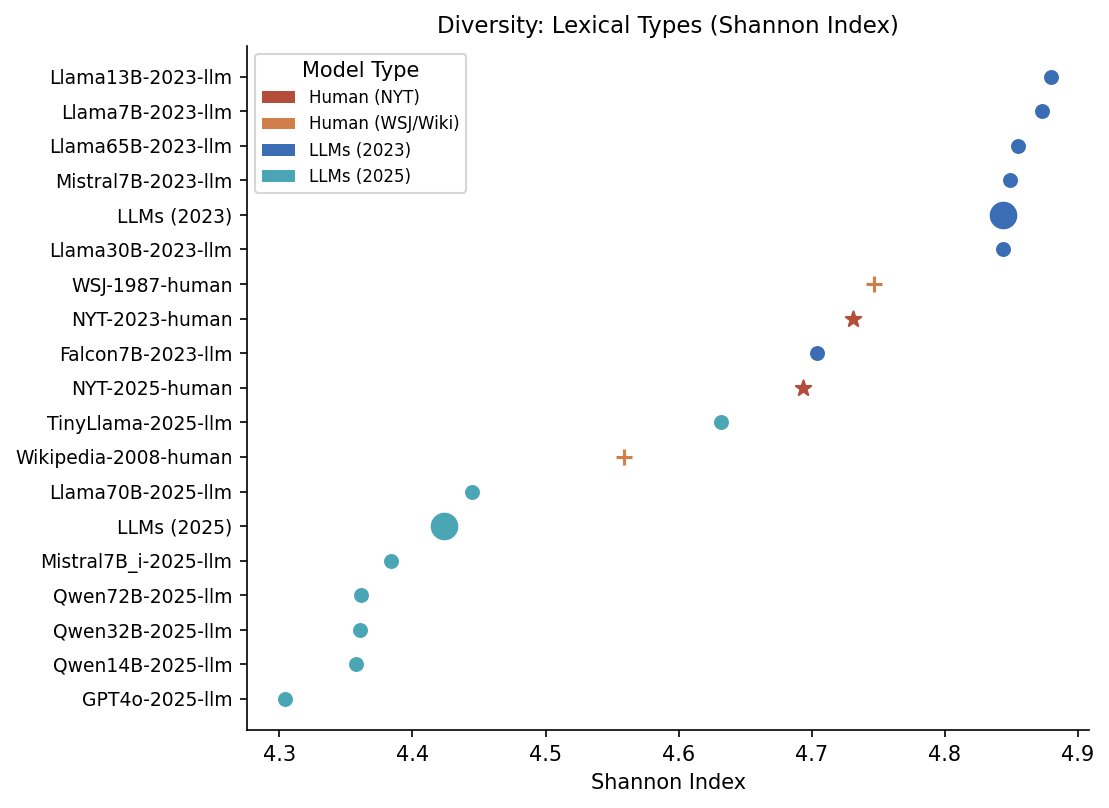}}}
                     \caption{Lexical type diversity measured using the Shannon Index. Higher values indicate a more varied distribution of lexical constructions. On the Y-axis, each point corresponds to a model name.}
                    \label{fig:erg-lex-type-shannon}
              \end{minipage}%
              \hfill
                   \begin{minipage}[t]{0.48\textwidth}
                    \centering
                    \vtop{\null\hbox{\includegraphics[width=\linewidth,clip]{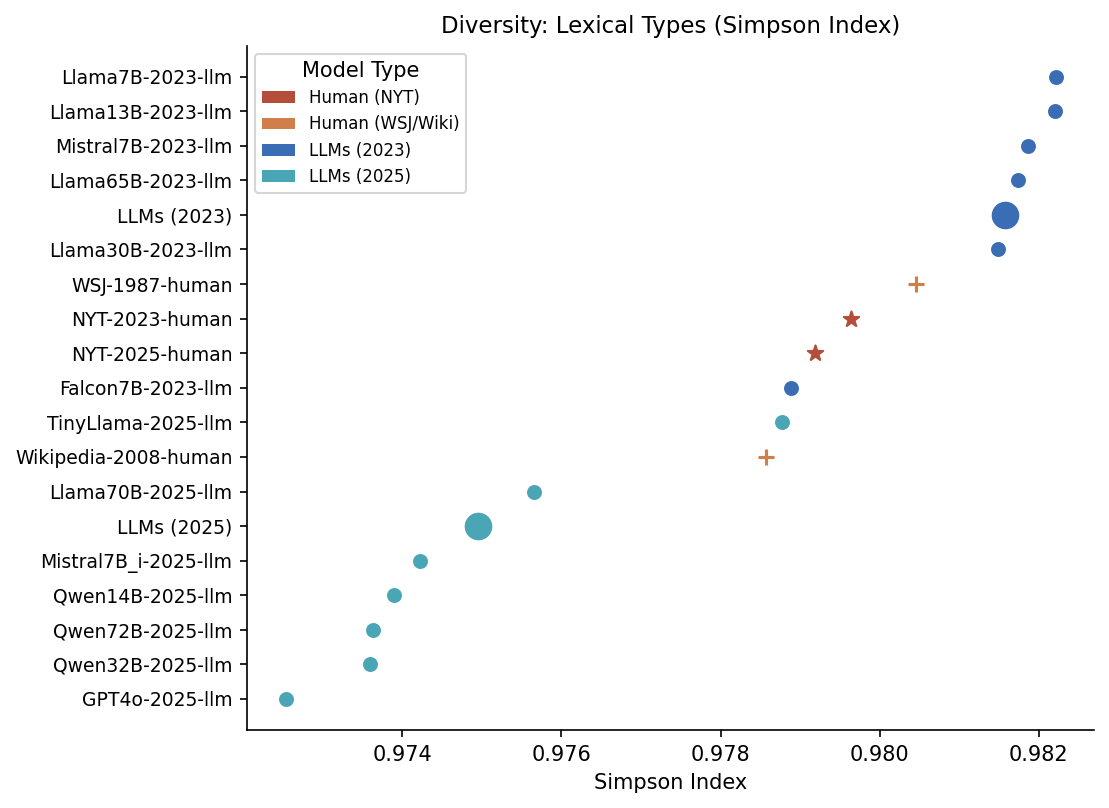}}}
                     \caption{Lexical type diversity measured using the Simpson Index. Higher values indicate a more varied distribution of lexical constructions. On the Y-axis, each point corresponds to a model name.}
                    \label{fig:erg-lex-type-simpson}
                \end{minipage}
            \end{figure*}
    
            Table \ref{tab:syntrank} 
            shows that, compared to LLM-generated texts, human-authored English news text makes frequent use of constructions that help bind text to concrete events, locations, and temporal frames. In particular, human-authored news texts show higher use of clause-embedding and attribution structures, a tendency that correlates with reportive verbs requiring subordinate clauses (such as `said' in `Critics said...', Table \ref{tab:lexrank}).

            Figure~\ref{fig:erg-const-shannon} shows that humans are clearly more diverse in their use of syntactic constructions than all LLMs.\footnote{Shannon indices were computed using maximum bootstrap = 10,000} In the case of the human texts, NYT-authored texts retrieved from 2025 are slightly less varied than the 2023 texts, but very close (seen also in terms of the Simpson index; Figure~\ref{fig:erg-const-simpson}).

            

            \subsection{Syntactic types: 2023 and 2025 LLMs}
            \label{sec:synt_evo}
            
                Table \ref{tab:llm_syntconst_side_by_side} gives some examples of the distributional differences between the LLMs from 2023 and  the LLMs from 2025. Notably, the newer 2025 LLMs are \textbf{not} more diverse than 2023 models; in fact, they form a \textbf{lower}-diversity band, despite being larger and trained on more recent data. This appears to be the result of the newer LLMs avoiding specifics such as names, dates, measures, etc., which influences not only lexical but also syntactic distribution. In this sense, older models were more like human writers.  
                
                In 2023, the constructions that contribute the most to the differences in diversity in comparison with 2025 are adjective-headed phrases and bare noun phrases (Table \ref{tab:llm_syntconst_side_by_side}). In contrast, the 2025 models favor constructions that involve  modification and coordination. These include noun phrases with modifiers, participial subordinate phrases, and coordinated noun structures. All of these add volume to the output but not necessarily content. Although bare noun phrases (common in the use of proper names) are present, they are less characteristic of the overall distribution, ranking 4th instead of 2nd (Table \ref{tab:llm_syntconst_side_by_side}). Overall, these results suggest that newer LLMs are more careful about outputting factually incorrect information, which, curiously, can be detected at the level of syntax.
                
            \subsection{Lexical types: Humans and LLMs}
            \label{sec:lex_types}
    
        

                Figure~\ref{fig:erg-lex-type-shannon} shows an interesting difference with respect to lexical types. While syntactically, human authors of English news were clearly the most diverse, when it comes to lexical types, LLM outputs from 2023 show the highest diversity, surpassing all other corpora. Human-authored English news texts sit in the middle, and LLMs from 2025 rank the lowest. This indicates that, despite being larger and trained on more recent corpora, the newer systems employ a lesser variety of vocabulary groupings characterized by certain syntactic behavior (an example of such a group would be mass nouns, or clause-embedding verbs, etc).  This surprising result calls for further investigation, possibly in the dimension of training paradigms, post-training and alignment. The same ranking is observed with Simpson indices (Figure~\ref{fig:erg-lex-type-simpson}). Humans, meanwhile, remain a stable reference, showing similar lexical type diversity as before, with a particularly distinctive use of proper names (Table \ref{tab:lexrank}).
                
    

            \subsection{Lexical types: 2023 and 2025 LLMs}
            \label{sec:lex_evo}
            
                The lexical-type ranking in Table \ref{tab:llm_lextypes_side_by_side} shows the change in the lexical types contributing the most to the diversity of the LLMs' distributions between 2023 and 2025. In 2023, the lexical types contributing the most to the diversity are utterance particles and personal pronouns. This suggests that the models' output is oriented toward conversational framing and speaker reference, with less emphasis on factual content. 
                By contrast, 2025 models have a distinctive distribution of the very `basic' lexical types: adjectives, common nouns, and transitive verbs. This may be a feature of a generic style that avoids specifics. 
                
         \subsection{Punctuation}
         \label{sec:other_ling}

\begin{figure*}[h!]
                \centering
                \begin{minipage}[h]{0.48\textwidth}
                    \centering
                    \includegraphics[width=\linewidth,clip]{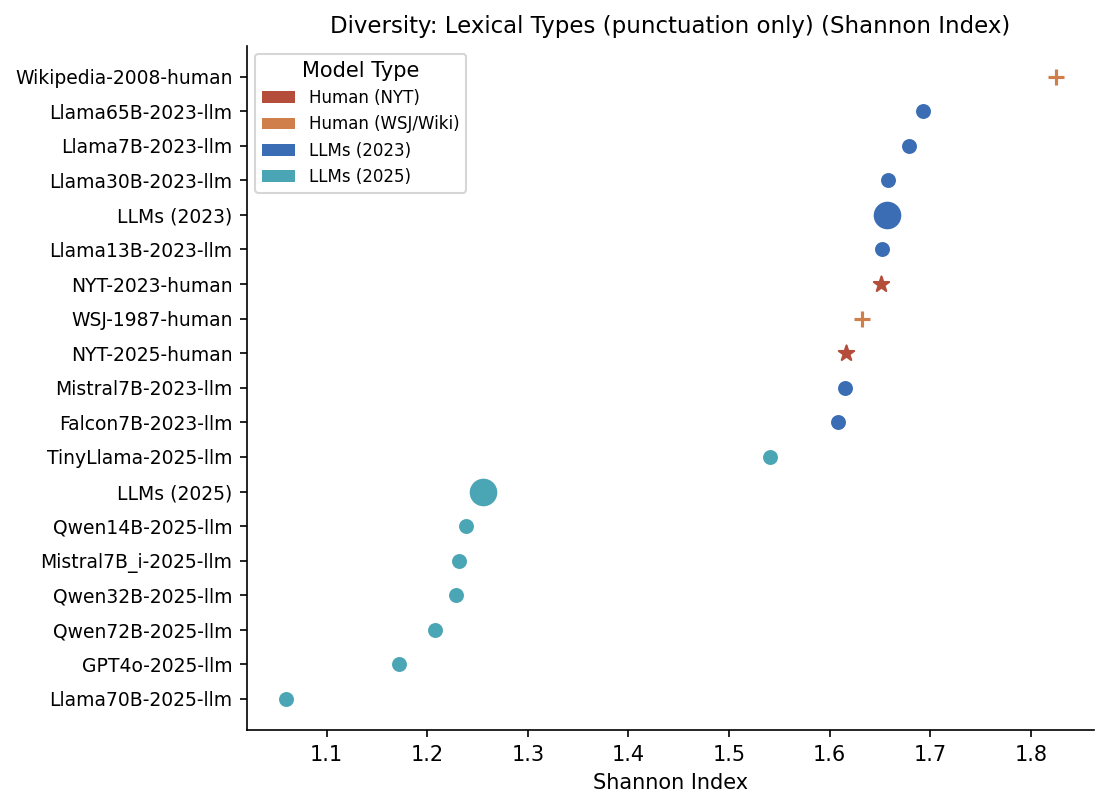}
                     \caption{Lexical type diversity measured using the Shannon Index considering only punctuation. On Y-axis, each point corresponds to a model name.}
                    \label{fig:erg-lex-type-shannon-punct}
                \end{minipage}
                \hfill
                \begin{minipage}[h]{0.48\textwidth}
                    \centering
                    \includegraphics[width=\linewidth,clip]{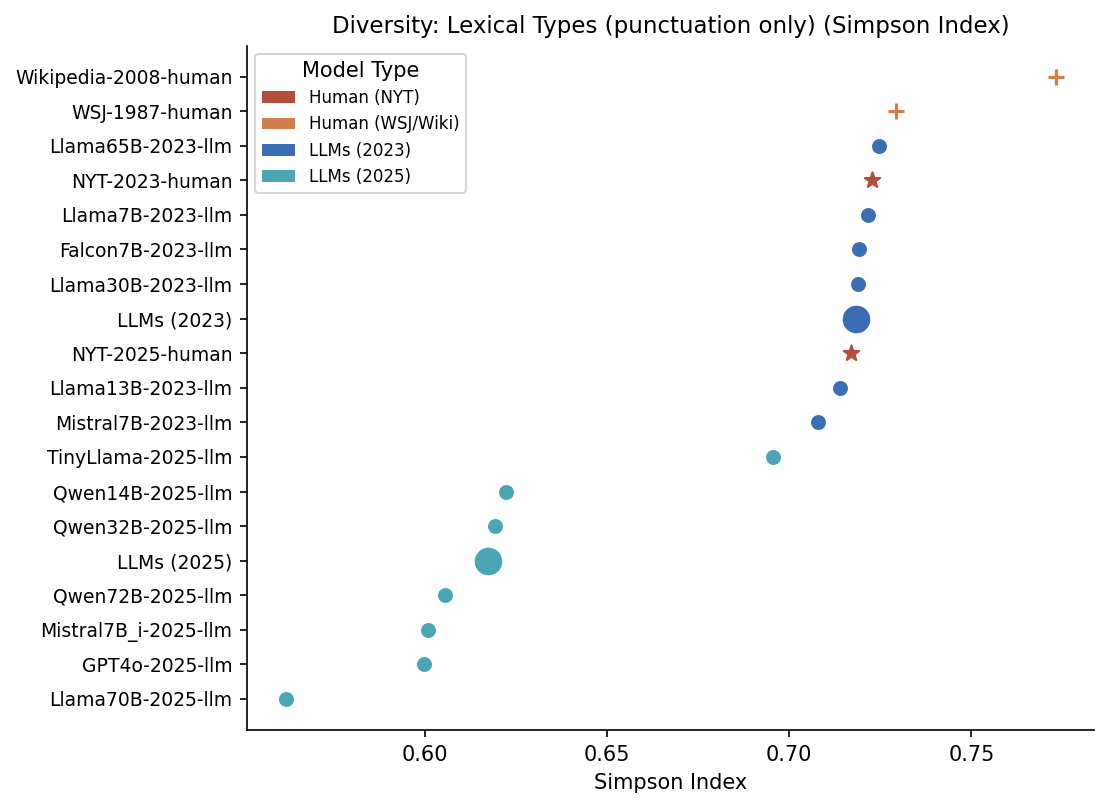}
                     \caption{Lexical type diversity measured using the Simpson Index considering only punctuation. On Y-axis, each point corresponds to a model name.}
                    \label{fig:erg-lex-type-simpson-punct}
                \end{minipage}
            \end{figure*}

\begin{figure*}[h!]
                \centering
                \begin{minipage}[h]{0.48\textwidth}
                    \centering
                    \includegraphics[width=\linewidth,clip]{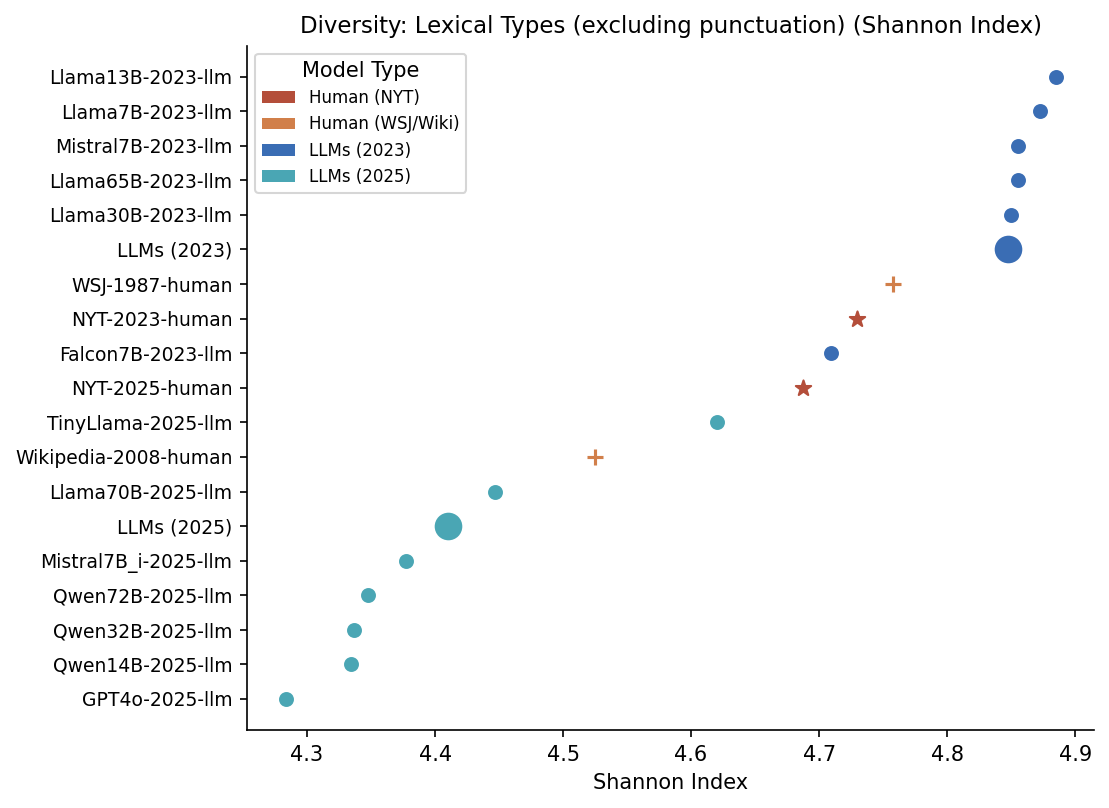}
                     \caption{Lexical type diversity measured using the Shannon Index excluding punctuation. On Y-axis, each point corresponds to a model name.}
                    \label{fig:erg-lex-type-shannon-xpunct}
                \end{minipage}
                \hfill
                \begin{minipage}[h]{0.48\textwidth}
                    \centering
                    \includegraphics[width=\linewidth,clip]{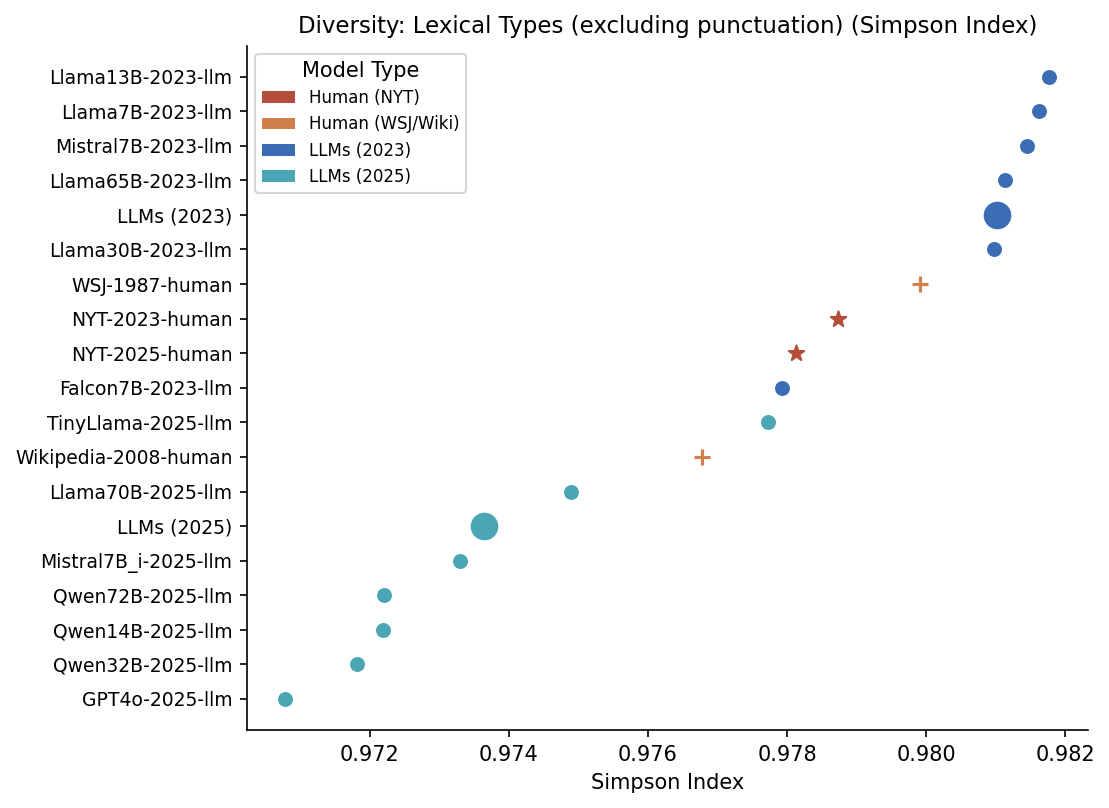}
                     \caption{Lexical type diversity measured using the Simpson Index excluding punctuation. On Y-axis, each point corresponds to a model name.}
                    \label{fig:erg-lex-type-simpson-xpunct}
                \end{minipage}
            \end{figure*}

            When we split the lexical types into punctuation and non-punctuation, in Figures \ref{fig:erg-lex-type-shannon-punct} and \ref{fig:erg-lex-type-shannon-xpunct} with Shannon index and Figures \ref{fig:erg-lex-type-simpson-punct} and \ref{fig:erg-lex-type-simpson-xpunct} with Simpson index, we see a very clear distinction: the 2025 models are much less diverse in their usage of punctuation. 
        
            Some anomalies are seen in GPT-4o: it generates very few em-dashes or semicolons, even though they are common in journalists' writing and they have been described as common in LLM-generated text~\citep{srivastava2025llms}.  It also generates very few quotations (correlating with its diminished use of the reportive verbs such as  `said'), whereas the rest of the models have shown no special behaviors like these.  For example, the WSJ has many sentences like this: \textit{``In Asia, as in Europe, a new order is taking shape,'' Mr. Baker said.}, but these are very rare in the output from GPT-4o. We hypothesize that this is a response to post-training aimed at reducing hallucinations and non-factual output.  Still, this is a significant difference with what we expect from newspaper text.

        \subsection{Text length and parsability}

        The length of the generated texts is also different between the 2023 and 2025 models. 
        More specifically, the 2025 models consistently generate longer sentences than both human-authored English news sentences and the sentences produced by 2023 models. In 2023, human sentences were about 10–20\% longer than LLM sentences. In contrast, newer models produce sentences that are 15–30\% longer than sentences written by human authors. The 2025 systems also reduce short sentences (1–15 tokens) by factors of 9 to 30 and cut non-sentence fragments by a factor of five. 
        Despite this, the ERG requires \textbf{fewer} resources (time and space) to parse the sentences generated by 2025 models. Despite generating longer sentences, newer LLMs do not appear to create more complex structures. Instead, they are easier for the ERG to parse (recall the ERG is a deterministic, exhaustive search chart parser, which can run out of time or space to parse a sentence). Human-authored English news sentences from both 2023 and 2025 reach about 94–95\% parse success, while LLM texts exceed 96\% and approach 99\% for the newest models. Even with substantial length increases, 2025 LLM sentences remain highly parsable, reflecting structural regularity that aligns with the best-understood and streamlined parts of the grammar, unlike human-authored English news text, which triggers phenomena that may be less understood or objectively challenging to parse. 
        This means that the
        that the text generated by the newer instruction-tuned LLMs is easier for the ERG to parse than both earlier LLM outputs and English human-authored news text. This  finding motivates future work exploring how specific grammatical structures correlate with parsability. More details about time and space required to parse each dataset are Appendix~\ref{sec:parsability}.

    \section{Conclusion}
    \label{sec:conclusion}
        In this work, we compared two generations of LLMs with two temporal samples of human-authored news writing, studying their syntactic structure within a formal linguistic framework. By applying diversity metrics from ecology and information theory to distributions of grammar constructions in LLM-generated and human-authored writing, we provided an interpretable view of how model-generated text changed over time relative to human language production. Our findings show that while English news text remains stable in both syntactic and lexical diversity, LLMs exhibit a shift: newer instruction-tuned systems produce text that is syntactically and lexically less diverse. 
 
        Despite producing longer sentences compared to human writers and to older LLMs, the 2025 instruction-tuned LLMs display reduced lexical variety and lower constructional diversity. The newer models also generate text that is easier to parse, suggesting increased predictability. Together, these trends indicate that the more recent instruction-tuned LLMs are constrained to a less expressive language space, yielding outputs that are more formulaic and less varied than English news text.

        Our results confirm with two independent frameworks\,---\,ecology diversity metrics combined with a linguistic-theoretic account of grammar\,---\,that larger and newer models are not closer to human linguistic behavior, at least in the domain of professional news writing. Instead, even though their fluency seems to have improved, they show a growing divergence in lexical and grammatical diversity from humans. Future work could examine how the instruction-tuning of the models affects their ability to generate more diverse texts, and if with the right procedures, we could enhance these linguistic capabilities that they seem to have traded off.

    \section*{Limitations}
        There are some limitations in terms of methodology based on the nature of the study. First, working with LLMs, which are non-deterministic models, introduces variability in the generated outputs, as results depend strongly on both the prompt design and the specific model used.
        
        Second, this study focuses primarily on NYT-style news articles, which do not fully represent the broader spectrum of writing styles.

        Third, the analysis of more recent LLMs is constrained by hardware limitations. Due to the large size of some of the models considered, quantization is applied during the inference phase when generating synthetic data. Specifically, we apply 4-bit quantization to the largest models (LLaMA 65B and Qwen 2.5 72B). In addition, LLaMA 65B is also evaluated under 8-bit quantization, while the remaining models are used without quantization.
        
        Finally, another limitation is the scarcity of large-scale HPSG grammars. Currently, only a few languages have wide-coverage implementations, and among them the English Resource Grammar is the only one extensive enough to parse roughly 94\% of news text. Consequently, the present study is necessarily restricted to English.

    \section*{Acknowledgments}
        
        We acknowledge grants GAP (PID2022-139308OA-I00) funded by MICIU/AEI/10.13039/501100011033/ and ERDF, EU; LATCHING (PID2023-147129OB-C21) funded by MICIU/AEI/10.13039/501100011033 and ERDF, EU; and TSI-100925-2023-1 funded by Ministry for Digital Transformation and Civil Service and ``NextGenerationEU'' PRTR; as well as funding by Xunta de Galicia (ED431C 2024/02). 
        
        CITIC, as a center accredited for excellence within the Galician University System and a member of the CIGUS Network, receives subsidies from the Department of Education, Science, Universities, and Vocational Training of the Xunta de Galicia. Additionally, it is co-financed by the EU through the  FEDER Galicia 2021-27 operational program (Ref. ED431G 2023/01).
            
        We have used ChatGPT and Gemini for minor copy-editing (e.g.\ thesaurus suggestions) and for visualization ideas. We have used GitHub copilot for code autocompletion and Claude Code for final refactoring. 
            
    \bibliography{llm-syntax}

@article{shypula2025evaluating,
  title={Evaluating the diversity and quality of llm generated content},
  author={Shypula, Alexander and Li, Shuo and Zhang, Botong and Padmakumar, Vishakh and Yin, Kayo and Bastani, Osbert},
  journal={arXiv preprint arXiv:2504.12522},
  year={2025}
}

@article{munoz2024contrasting,
  title={Contrasting linguistic patterns in human and {LLM}-generated news text},
  author={Muñoz-Ortiz, Alberto and Gómez-Rodríguez, Carlos and Vilares, David},
  journal={Artificial Intelligence Review},
  volume={57},
  number={10},
  pages={265},
  year={2024},
  publisher={Springer}
}

@inproceedings{zamaraeva2025comparing,
    title = "Comparing {LLM}-generated and human-authored news text using formal syntactic theory",
    author = "Zamaraeva, Olga  and
      Flickinger, Dan  and
      Bond, Francis  and
      G{\'o}mez-Rodr{\'i}guez, Carlos",
    editor = "Che, Wanxiang  and
      Nabende, Joyce  and
      Shutova, Ekaterina  and
      Pilehvar, Mohammad Taher",
    booktitle = "Proceedings of the 63rd Annual Meeting of the Association for Computational Linguistics (Volume 1: Long Papers)",
    month = jul,
    year = "2025",
    address = "Vienna, Austria",
    publisher = "Association for Computational Linguistics",
    url = "https://aclanthology.org/2025.acl-long.443/",
    doi = "10.18653/v1/2025.acl-long.443",
    pages = "9041--9060",
    ISBN = "979-8-89176-251-0"
}

@article{reinhart2025llms,
  title={Do LLMs write like humans? Variation in grammatical and rhetorical styles},
  author={Reinhart, Alex and Markey, Ben and Laudenbach, Michael and Pantusen, Kachatad and Yurko, Ronald and Weinberg, Gordon and Brown, David West},
  journal={{Proceedings of the National Academy of Sciences}},
  volume={122},
  number={8},
  pages={e2422455122},
  year={2025},
  publisher={National Academy of Sciences}
}

@article{moon2025homogenizing,
  title={Homogenizing Effect of Large Language Models (LLMs) on Creative Diversity: An Empirical Comparison of Human and ChatGPT Writing},
  author={Moon, Kibum and Green, Adam E and Kushlev, Kostadin},
  journal={Computers in Human Behavior: Artificial Humans},
  pages={100207},
  year={2025},
  publisher={Elsevier}
}

@inproceedings{rashid2024humanizing,
  title={Humanizing AI in Education: A Readability Comparison of LLM and Human-Created Educational Content},
  author={Rashid, Md Mamunur and Atilgan, Nilsu and Dobres, Jonathan and Day, Stephanie and Penkova, Veronika and K{\"u}{\c{c}}{\"u}k, Mert and Clapp, Steven R and Sawyer, Ben D},
  booktitle={Proceedings of the Human Factors and Ergonomics Society Annual Meeting},
  volume={68},
  number={1},
  pages={596--603},
  year={2024},
  organization={SAGE Publications Sage CA: Los Angeles, CA}
}

@book{biber1991variation,
  title={Variation across speech and writing},
  author={Biber, Douglas},
  year={1991},
  publisher={Cambridge University Press}
}

@article{flickinger2000building,
  title={On building a more efficient grammar by exploiting types},
  author={Flickinger, Dan},
  journal={Natural Language Engineering},
  volume={6},
  number={01},
  pages={15--28},
  year={2000},
  publisher={Cambridge Univ Press}
}

@incollection{Flickinger:11,
  author = {Flickinger, Dan},
  title = {Accuracy v.\ Robustness in Grammar Engineering},
  booktitle = {Language from a Cognitive Perspective: Grammar, Usage and Processing},
  editor = {Bender, Emily M. and Arnold, Jennifer E.},
  address = {Stanford, CA},
  publisher = {CSLI Publications},
  year = 2011,
  pages = {31--50}
}

@book{Pol:Sag:94,
    author = {Carl Pollard and Ivan A. Sag},
    title = {{H}ead-{D}riven {P}hrase {S}tructure {G}rammar},
    year= 1994,
    publisher = {The University of Chicago Press and CSLI Publications},
    series = {Studies in Contemporary Linguistics},
    address = {Chicago, IL and Stanford, CA}
}

@inproceedings{nivre2016universal,
  title={Universal dependencies v1: A multilingual treebank collection},
  author={Nivre, Joakim and De Marneffe, Marie-Catherine and Ginter, Filip and Goldberg, Yoav and Hajic, Jan and Manning, Christopher D and McDonald, Ryan and Petrov, Slav and Pyysalo, Sampo and Silveira, Natalia and Tsarfaty, Reut},
  booktitle={Proceedings of the Tenth International Conference on Language Resources and Evaluation (LREC'16)},
  pages={1659--1666},
  year={2016}
}

@article{stamatatos2009survey,
  title={A survey of modern authorship attribution methods},
  author={Stamatatos, Efstathios},
  journal={Journal of the American Society for Information Science and Technology},
  volume={60},
  number={3},
  pages={538--556},
  year={2009},
  publisher={Wiley},
  doi={10.1002/asi.21001},
  url={https://www.researchgate.net/publication/220435062}
}

@inproceedings{Ben:Fli:Oep:02,
  author = {Bender, Emily M. and Flickinger, Dan and Oepen, Stephan},
  title = {The {G}rammar {M}atrix: {A}n open-source starter-kit for the rapid development of cross-linguistically consistent broad-coverage precision grammars},
  booktitle = {Proceedings of the {W}orkshop on Grammar Engineering and Evaluation at the 19th {I}nternational {C}onference on {C}omputational {L}inguistics},
  editor = {Carroll, John and Oostdijk, Nelleke and Sutcliffe, Richard},
  address = {Taipei},
  pages = {8--14},
  year= {2002}
}

@misc{padmakumar2024doeswritinglanguagemodels,
      title={Does Writing with Language Models Reduce Content Diversity?}, 
      author={Vishakh Padmakumar and He He},
      year={2024},
      eprint={2309.05196},
      archivePrefix={arXiv},
      primaryClass={cs.CL},
      url={https://arxiv.org/abs/2309.05196}, 
}

@misc{kirk2024understandingeffectsrlhfllm,
      title={Understanding the Effects of RLHF on LLM Generalisation and Diversity}, 
      author={Robert Kirk and Ishita Mediratta and Christoforos Nalmpantis and Jelena Luketina and Eric Hambro and Edward Grefenstette and Roberta Raileanu},
      year={2024},
      eprint={2310.06452},
      archivePrefix={arXiv},
      primaryClass={cs.LG},
      url={https://arxiv.org/abs/2310.06452}, 
}

@article{https://doi.org/10.1046/j.1466-822X.2003.00015.x,
author = {Spellerberg, Ian F. and Fedor, Peter J.},
title = {A tribute to Claude Shannon (1916–2001) and a plea for more rigorous use of species richness, species diversity and the ‘Shannon–Wiener’ Index},
journal = {Global Ecology and Biogeography},
volume = {12},
number = {3},
pages = {177-179},
doi = {https://doi.org/10.1046/j.1466-822X.2003.00015.x},
url = {https://onlinelibrary.wiley.com/doi/abs/10.1046/j.1466-822X.2003.00015.x},
eprint = {https://onlinelibrary.wiley.com/doi/pdf/10.1046/j.1466-822X.2003.00015.x},
year = {2003}
}

@article{Simpson1949,
  author  = {E. H. Simpson},
  title   = {Measurement of diversity},
  journal = {Nature},
  volume  = {163},
  pages   = {688},
  year    = {1949},
  doi     = {10.1038/163688a0}
}

@book{Magurran2004,
  author    = {Anne E. Magurran},
  title     = {Measuring Biological Diversity},
  publisher = {Blackwell Publishing},
  address   = {Oxford},
  year      = {2004},
  isbn      = {978-0-632-05633-0}
}

@online{srivastava2025llms,
  author = {Srivastava, Rajesh},
  title = {How {LLM}s Turned the Em Dash (—) Into a Villain - {T}echnical Nuances},
  year = {2025},
  month = {aug},
  url = {https://medium.com/@raj-srivastava/how-llms-turned-the-em-dash-into-a-villain-technical-nuances-b564857adc3b},
  urldate = {2025-10-07}
}

@inproceedings{NEURIPS2022_b1efde53,
 author = {Ouyang, Long and Wu, Jeffrey and Jiang, Xu and Almeida, Diogo and Wainwright, Carroll and Mishkin, Pamela and Zhang, Chong and Agarwal, Sandhini and Slama, Katarina and Ray, Alex and Schulman, John and Hilton, Jacob and Kelton, Fraser and Miller, Luke and Simens, Maddie and Askell, Amanda and Welinder, Peter and Christiano, Paul F and Leike, Jan and Lowe, Ryan},
 booktitle = {Advances in Neural Information Processing Systems},
 editor = {S. Koyejo and S. Mohamed and A. Agarwal and D. Belgrave and K. Cho and A. Oh},
 pages = {27730--27744},
 publisher = {Curran Associates, Inc.},
 title = {Training language models to follow instructions with human feedback},
 url = {https://proceedings.neurips.cc/paper_files/paper/2022/file/b1efde53be364a73914f58805a001731-Paper-Conference.pdf},
 volume = {35},
 year = {2022}
}
    
    \appendix
    
    \section*{Appendices}
    \label{appendices}
    
        \section{Parsability}
        \label{sec:parsability}

            Tables~\ref{tab:unbinned} and~\ref{tab:binned} show that newer instruction-tuned LLMs produce texts that are easier to parse than both human-authored English news text and earlier LLM outputs.


            \begin{table*}
                \begin{center}
                    \begin{tabular}{|l|r|r|r|r|r|r|r|r|}
                        \hline
                        Profile              & Items & Parsed & Length & Short & Frgmt & Time & Space & $>$Limit\\
                        2023 (RAM limit 21G) &       &   \%   & toks/S &   \%  &   \%  & sec/S & Gb/S &  \%   \\
                        \hline                                                                      
                        \emph{nyt-2023-human}	     & \emph{26092} & \emph{93.4}   & \emph{22.33}  &  \emph{33}   &  \emph{13}  &  \emph{11.5} & \emph{2.3}  &  \emph{4.3}  \\
                        falcon07-2023-llm    & 27769 & 97.7   & 18.62  &  37   &  10  &   4.4 & 0.9  &  1.1  \\
                        llama07-2023-llm     & 37825 & 96.4   & 19.42  &  35   &  12  &   6.0 & 1.2  &  1.8  \\
                        llama13-2023-llm     & 37800 & 97.1   & 18.60  &  38   &  13  &   5.1 & 1.0  &  1.4  \\
                        llama30-2023-llm     & 37568 & 96.9   & 18.17  &  39   &  12  &   4.9 & 1.0  &  1.3  \\
                        llama65-2023-llm     & 38107 & 96.4   & 18.76  &  37   &  12  &   5.7 & 1.1  &  1.7  \\
                        mistral7b-2023-llm   & 35086 & 97.3   & 18.36  &  38   &  13  &   4.7 & 0.9  &  1.1  \\
                        \hline                                                                      
                        2025 (RAM limit 31G) & & & & & & & & \\                                     
                        \hline                                                                      
                        \emph{nyt-2025-human}       & \emph{24053} & \emph{94.8}   & \emph{22.19}  &  \emph{32}   &  \emph{12}  &  \emph{12.2} & \emph{2.4}  &  \emph{2.6}  \\
                        qwen14-2025-llm      & 26498 & 98.2   & 25.63  &   4   &   2  &   7.2 & 1.6  &  0.6  \\
                        qwen32-2025-llm	     & 34892 & 98.4   & 25.62  &   3   &   2  &   6.7 & 1.3  &  0.5  \\
                        qwen72-2025-llm      & 34614 & 98.5   & 26.05  &   2   &   2  &   6.0 & 1.2  &  0.3  \\
                        llama70-2025-llm     & 39306 & 97.9   & 29.87  &   1   &   2  &   9.5 & 2.0  &  0.9  \\
                        gpt4o-2025-llm       & 50544 & 98.5   & 25.99  &   1   &   2  &   4.6 & 1.0  &  0.2  \\
                        mistral7i-2025-llm   & 39708 & 98.3   & 25.87  &   4   &   2  &   7.4 & 1.4  &  0.6  \\
                        \hline
                    \end{tabular}
                \end{center}
                \caption{Parsing statistics for human-authored (in italics) and LLM-generated news datasets from 2023 and 2025. Each row reports the percentage of sentences successfully parsed by the ERG, average sentence length in tokens, proportion of short sentences ($\le 15$ tokens), proportion of fragments, mean CPU time and memory per sentence, and proportion of sentences exceeding resource limits.}
                \label{tab:unbinned}
            \end{table*}

            \begin{table*}
                \begin{center}
                    \begin{tabular}{|l|r|r|r|r||r|r|r|r|}
                        \hline
                        Profile & \multicolumn{4}{|c|}{Time (CPU-seconds/sent)} & \multicolumn{4}{|c|}{Space (Gbytes/sent)} \\
                        (length in tokens) & 31-35 & 36-40 & 41-45 & 46-50 & 31-35 & 36-40 & 41-45 & 46-50 \\
                        \hline
                        \emph{nyt-2023-human}	   &    \emph{13} &    \emph{28} &    \emph{48} &    \emph{67} &   \emph{2.6} &   \emph{5.2} &   \emph{9.1} &  \emph{13.1} \\
                        falcon07-2023-llm  &    11 &    24 &    44 &    68 &   2.2 &   4.5 &   8.3 &  13.6 \\
                        llama07-2023-llm   &    14 &    26 &    50 &    68 &   2.7 &   5.0 &   9.5 &  13.4 \\
                        llama13-2023-llm   &    13 &    28 &    50 &    65 &   2.6 &   5.3 &   9.5 &  13.3 \\
                        llama30-2023-llm   &    14 &    28 &    51 &    69 &   2.6 &   5.3 &   9.4 &  13.3 \\
                        llama65-2023-llm   &    14 &    30 &    50 &    67 &   2.7 &   5.5 &   9.4 &  13.1 \\
                        mistral7b-2023-llm &    13 &    27 &    48 &    73 &   2.5 &   5.1 &   8.8 &  14.1 \\
                        \hline					                       
                        \emph{nyt-2025-human}     &    \emph{13} &    \emph{30} &    \emph{52} &    \emph{84} &   \emph{2.7} &   \emph{5.9} &  \emph{10.1} &  \emph{16.2} \\
                        qwen14-2025-llm    &    10 &    25 &    49 &    77 &   2.3 &   5.1 &   9.8 &  15.3 \\
                        qwen32-2025-llm	   &    10 &    24 &    50 &    78 &   2.0 &   4.4 &   8.7 &  14.0 \\
                        qwen72-2025-llm    &     9 &    21 &    42 &    67 &   1.8 &   4.0 &   7.6 &  12.1 \\
                        llama70-2025-llm   &     8 &    17 &    38 &    61 &   1.7 &   3.5 &   7.3 &  11.7 \\
                        gpt4o-2025-llm     &     8 &    17 &    36 &    54 &   1.6 &   3.3 &   6.5 &   9.8 \\
                        mistral7i-2025-llm &    10 &    25 &    50 &    81 &   2.0 &   4.5 &   8.7 &  14.0 \\
                        \hline
                    \end{tabular}
                \end{center}
                \caption{Average parsing cost by sentence-length bin for human (in italics) and LLM-generated texts.
                CPU time and memory consumption per sentence (in seconds and GB, respectively) for sentences binned by length (31–50 tokens).}
                \label{tab:binned}
            \end{table*}

\end{document}